\documentclass{article}
\pdfoutput=1

\usepackage[final,nonatbib]{nips_2018}
\usepackage[numbers]{natbib}

\usepackage[utf8]{inputenc} 
\usepackage[T1]{fontenc}    
\usepackage{hyperref}       
\usepackage{url}            
\usepackage{booktabs}       
\usepackage{amsfonts}       
\usepackage{nicefrac}       
\usepackage{microtype}      
\usepackage[capitalise]{cleveref}
\usepackage{graphicx}
\usepackage{xcolor}
\usepackage{amsmath}
\usepackage{subcaption}
\usepackage[english]{babel}

\DeclareMathOperator*{\argmax}{arg\,max}

\newcommand{\nipsfairparauthor}[2] {#1\\
  Facebook AI Research\\
  \texttt{#2}
}

\title{High-Level Strategy Selection under Partial Observability in StarCraft: Brood War}

\author{
\nipsfairparauthor{Jonas Gehring, Da Ju, Vegard Mella, Daniel Gant, Nicolas Usunier, Gabriel Synnaeve}{\{jgehring,\!daju,\!vegardmella,\!danielgant,\!usunier,\!gab\}@fb.com}
}

\begin{document}

\maketitle

\begin{abstract}
We consider the problem of high-level strategy selection in the adversarial setting of real-time strategy games from a reinforcement learning perspective, where taking an action corresponds to switching to the respective strategy.
Here, a good strategy successfully counters the opponent's current and possible future strategies which can only be estimated using partial observations.
We investigate whether we can utilize the full game state information during training time (in the form of an auxiliary prediction task) to increase performance.
Experiments carried out within a StarCraft\textsuperscript{\textregistered}: Brood War\textsuperscript{\textregistered}\footnote{StarCraft is a trademark or registered trademark of Blizzard Entertainment, Inc., in the U.S. and/or other countries.  Nothing in this paper should be construed as approval, endorsement, or sponsorship by Blizzard Entertainment, Inc.} bot against strong community bots show substantial win rate improvements over a fixed-strategy baseline and encouraging results when learning with the auxiliary task.
\end{abstract}

\section{Introduction}

Incomplete information, large state and action spaces, complex and stochastic but closed-world dynamics make real-time strategy (RTS) games such as \emph{StarCraft: Brood War} or \emph{DoTA 2} games an interesting test-bed for search, planning and reinforcement learning algorithms \cite{ontanon2013survey,vinyals2017starcraft}.
The ``fog of war'' is a fundamental aspect of RTS gameplay: players only observe the immediate surroundings of the units they control.
Hence, it is crucial to make good estimates about the strategy and positioning of an opponent and to produce matching counter-strategies in order to win a game.

In this work, we consider the selection between fixed high-level strategies (we will use the term \emph{build orders}; see also \cref{sec:actions}) for a \emph{StarCraft: Brood War} bot\footnote{We perform experiments with CherryPi (\url{https://torchcraft.github.io/TorchCraftAI})}, dependent on the current observable game state.
A build order consists of a rule set targeting specific unit compositions, decisions to expand to multiple bases and a global decision on whether to initiate attacks against the opponent or not. The bot we use in our experiments has 25 build orders to choose from. Learning strategic decisions for \emph{StarCraft} has previously been addressed in \cite{justesen2017learning} with a focus on learning individual commands from human replays, while we select among predefined strategies with reinforcement learning and evaluate against high-level competitive bots.

For this selection task, it is key to infer the strategy and actions of the opponent player from limited observations.
While it is possible to tackle hidden state estimation separately (e.g. \cite{lin2018forward} in the context of \emph{StarCraft}) and to provide a model with these estimates, we instead opt to perform estimation as an auxiliary prediction task alongside the default training objective.
Auxiliary losses (or multi-task learning) are well-known in neural-network based supervised learning~\cite{ando2005framework,zhang2016augmenting} and have recently found application in reinforcement learning tasks such as navigation (\cite{mirowski2016learning} employ auxiliary depth and loop closure prediction) and FPS game playing (\cite{dosovitskiy2016learning} predict future low-dimensional measurements; \cite{lample2017playing}~predict symbolic game features).
A common motivation in these works is to enable faster or more data-efficient learning of robust representations which facilitate mastering the actual control task at hand.
While we share this inspiration, our auxiliary task concerns present but hidden information.





\section{Approach}

Every five seconds of game time, we provide our model with a global, non-spatial representation of the current observation.
The features contain observed unit counts for both players (only partially observed for the opponent), our resources and technologies as well as game time and static game information such as the opponent faction. We define two settings for featurization: \emph{visible} only counts units currently visible, whereas \emph{memory} uses hard-coded rules to keep track of enemy units that were seen before but are currently hidden, as commonly done in \emph{StarCraft} bots. 
The model learns the value $Q(o, a)$ of switching to build-order $a$ given the observation $o$. It uses an LSTM encoder with 2048 cells followed by a linear layer with as many sigmoid outputs as build orders, and is trained with the win/lose outcome of the game as target. The auxilary task is dealt with by another branch of the network, taking the LSTM encoding as input. It consists of three fully connected layers of 256 hidden units with as many outputs as unit types. It  predicts the (nomalized) unit counts of the opponent by minimizing the Huber loss (the true opponent unit counts are available at training time\footnote{We activate BWAPI's \texttt{CompleteMapInformation} cheat flag for training games.}).
We train our models in two stages (see \cref{sec:features} and \ref{sec:training} for more details on the features and training):

\textbf{Off-policy Initialization:}
Offline training data is collected by performing random build order switches during a game. Specifically, we start by selecting an initial build order and perform a random switch every 8, 10 or 13 minutes on average (interval randomly selected for each game).
We produced a corpus of 2.8M games with 3.3M switches used as training data points for the Q-function. 

\textbf{On-policy Refinement:} 
We play games as in evaluation mode (\cref{sec:training}), selecting the build order according to the trained Q-function.
As before, we perform one random switch within a sampled average time interval and keep the selected build order for $T \sim \mathcal{U}(2,13)$ minutes.
Afterwards, we fall back to following the current Q-function.




\section{Results}
\label{sec:results}

\begin{table}
  \caption{Win rates for different setups after on-policy refinement.
  Evaluations are done on the bots in the training set (see \cref{tbl:bots}) and two bots that have not been seen during training.}
  \label{tbl:results}
  \begin{small}
  \begin{center}
  \begin{tabular}{llllll}
    \toprule
    Model & Unit Obs. & Loss & \multicolumn{3}{c}{Win Rate (std. dev)} \\
    \cmidrule(r){4-6}
    & & & Training bots & Locutus-20181007 & McRave-51e49b0 \\
    \midrule
    Control & - & - & 0.793 (0.010) & 0.387 (0.036) &  0.556 (0.032) \\
    \midrule
    LSTM & Visible & Value & 0.878 (0.004) & 0.635 (0.031) & 0.706 (0.048) \\ 
         & Visible & Value+Aux. & 0.879 (0.001) & 0.579 (0.065) & 0.723 (0.063) \\
         & Memory & Value & 0.886 (<0.001) & 0.587 (0.031) & 0.693 (0.077) \\
         & Memory & Value+Aux. & 0.888 (0.004) & 0.600 (0.027) & 0.725 (0.039) \\
    \bottomrule
  \end{tabular}
  \end{center}
  \end{small}
\end{table}




Our evaluation protocol is described in~\cref{sec:training}. 
\cref{tbl:results} compares win rates for a control run (without build order switching) and trained models when playing against the training set opponents and two held-out bot versions.
The training set win rate is the average of all per-bot win rates to account for the varying number of build orders per opponent faction.
Standard deviations are computed on averages for the first, second and third set of games per map iteration.

\begin{figure}[h]
\centering
\begin{subfigure}{.5\textwidth}
  \centering
  \includegraphics[width=\linewidth]{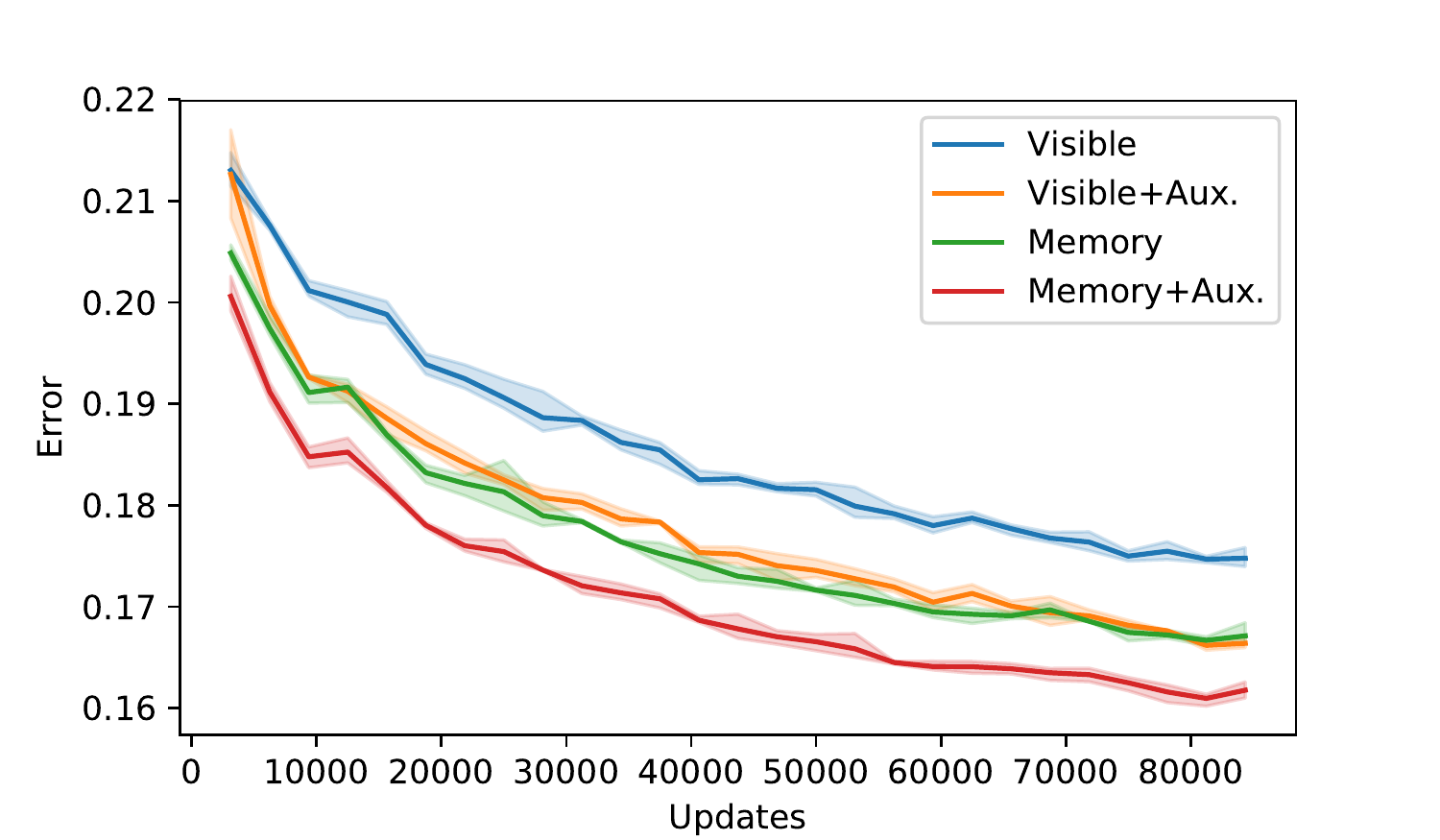}
  \caption{Q-value prediction errors}
  \label{fig:valid1}
\end{subfigure}%
\begin{subfigure}{.5\textwidth}
  \centering
  \includegraphics[width=\linewidth]{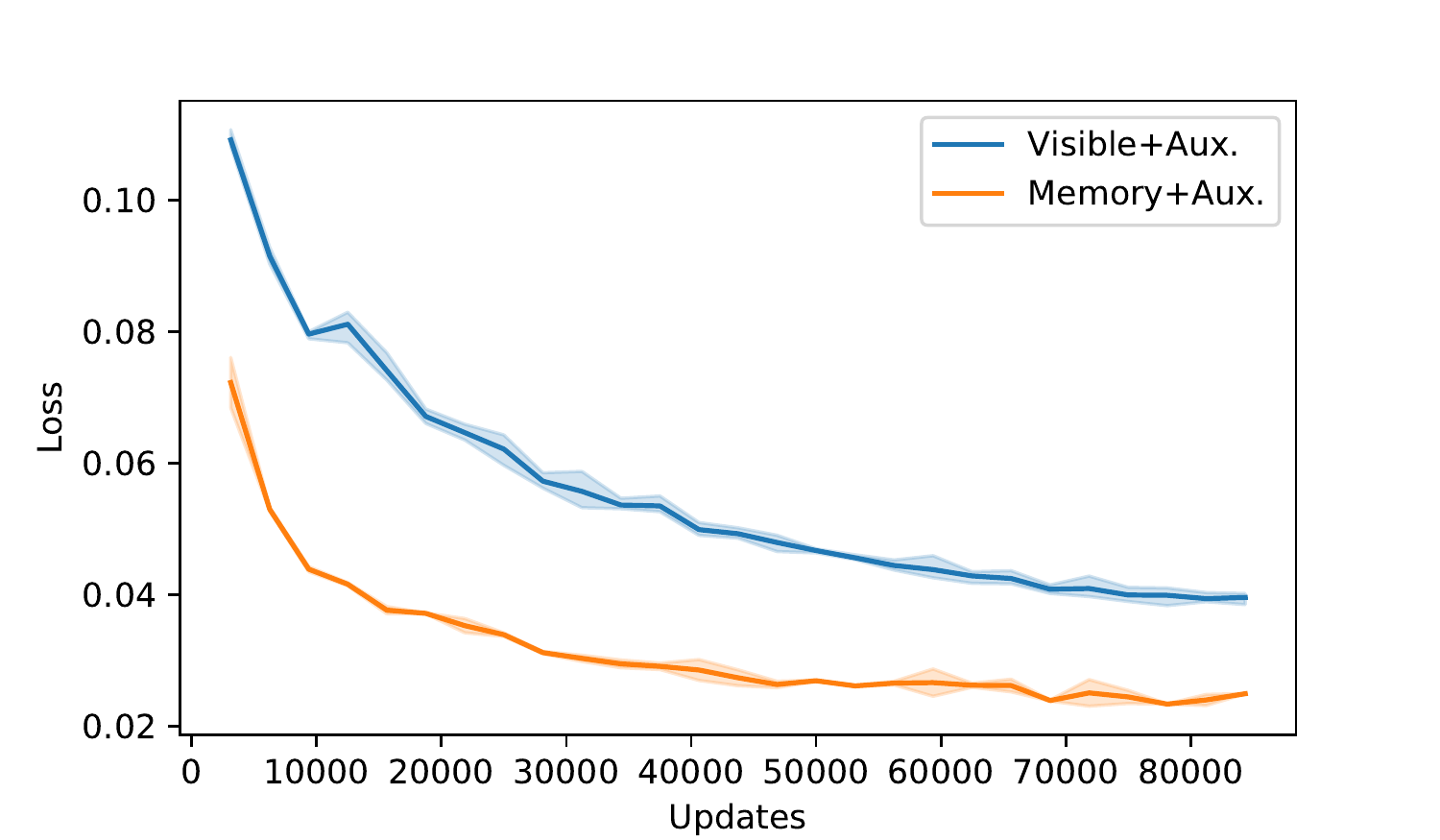}
  \caption{Auxiliary task Huber loss}
  \label{fig:valid2}
\end{subfigure}
\caption{Performance during off-policy initialization on a held-out in-domain validation set for Q-value prediction on build order switching points (errors) and the auxiliary task (Huber loss).
Data points are averaged over 3 runs each.
}
\label{fig:valid}
\end{figure}

\begin{figure}[h]
\centering
\begin{subfigure}{.5\textwidth}
  \centering
  \includegraphics[width=\linewidth]{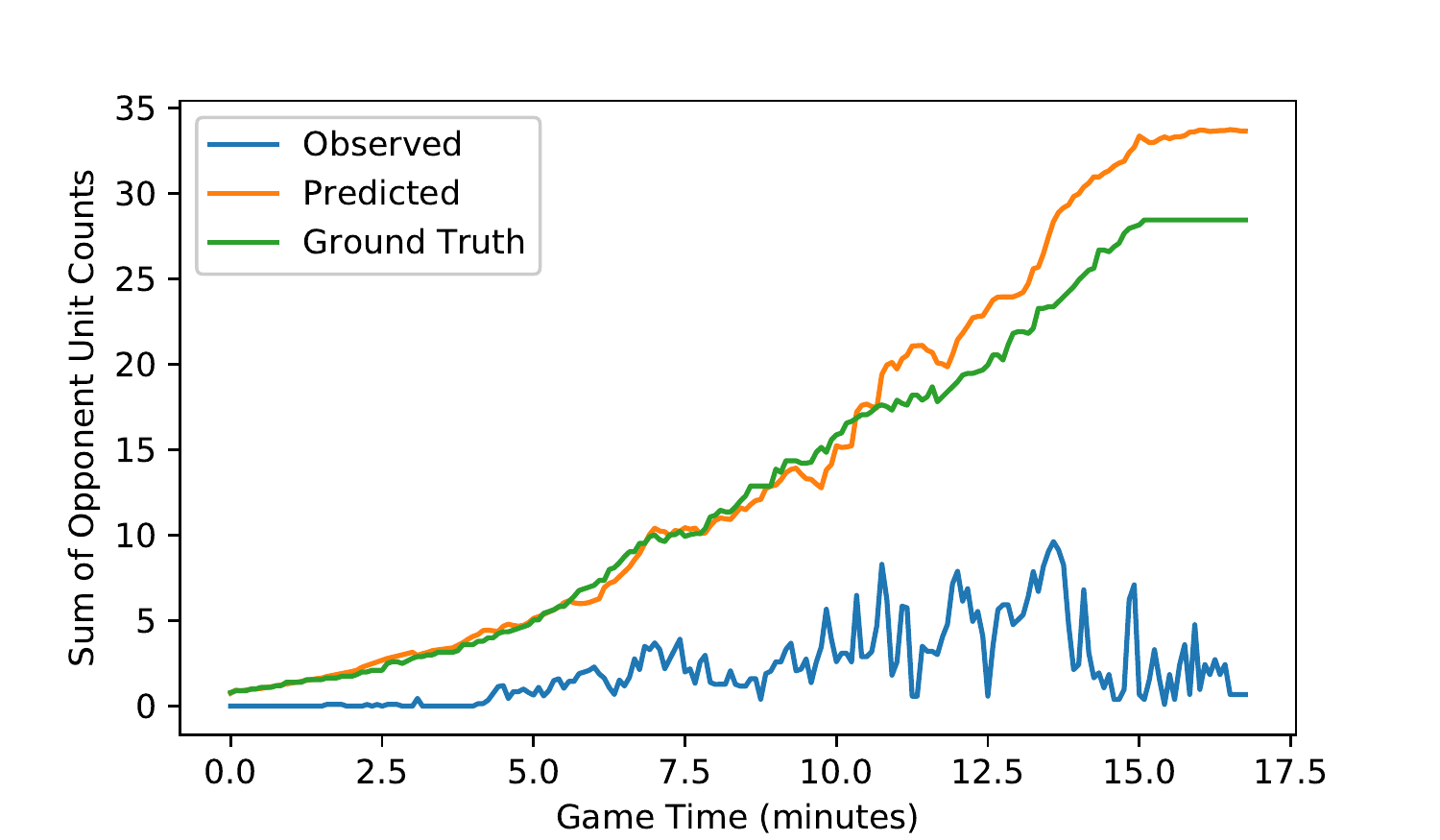}
  \caption{vs. Locutus-2018107}
  \label{fig:defogcounts1}
\end{subfigure}%
\begin{subfigure}{.5\textwidth}
  \centering
  \includegraphics[width=\linewidth]{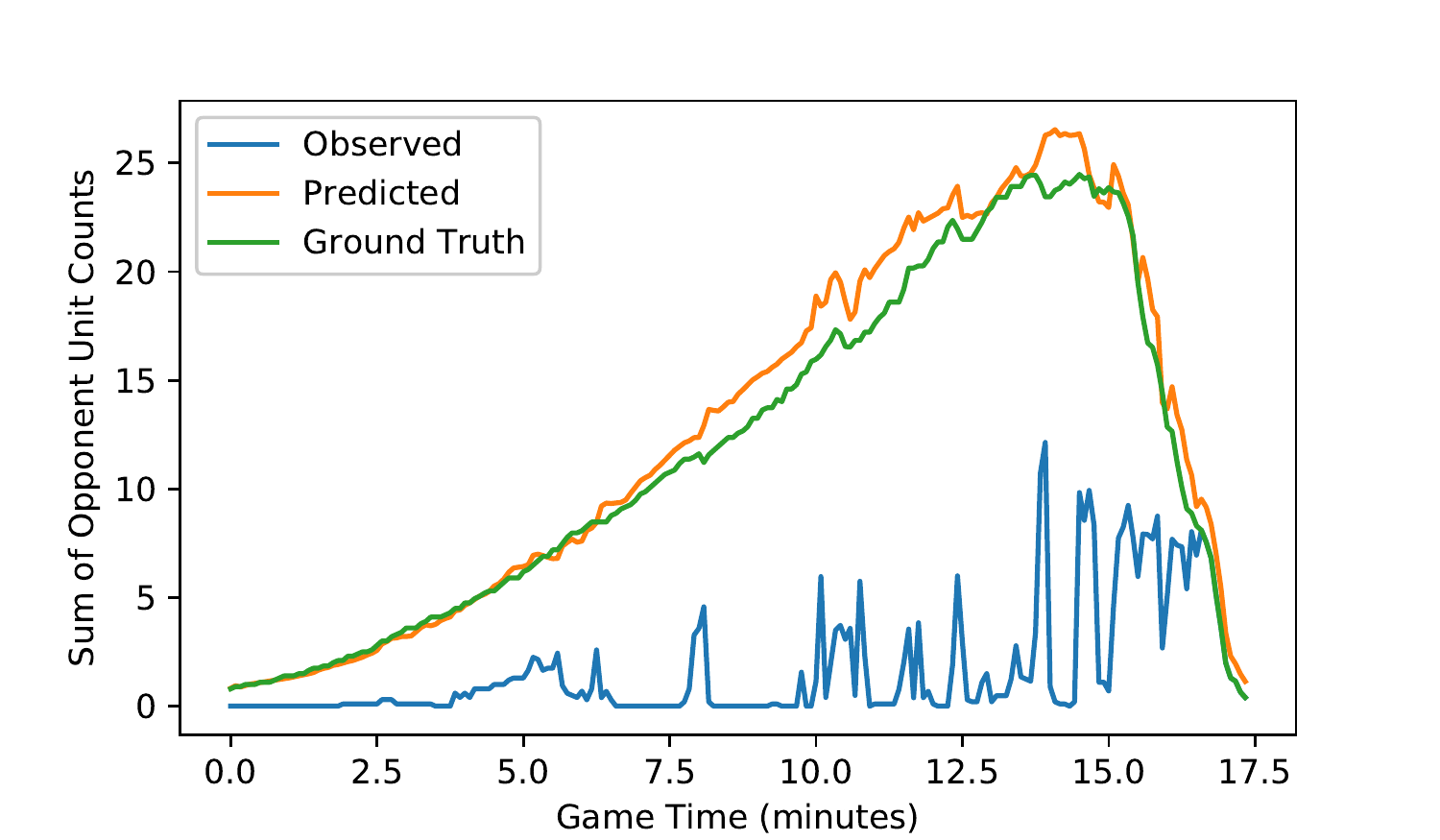}
  \caption{vs. McRave-51e49b0}
  \label{fig:defogcounts2}
\end{subfigure}
\caption{Observed (visible only), predicted and ground truth opponent unit counts against two bots not seen during training.
The y-axis represents a sum of the respective counts, individually normalized as described in~\cref{sec:features}.
In both cases, the unit count prediction closely follows the ground truth.
\cref{fig:defogcounts1} depicts a loss while the game in \cref{fig:defogcounts2} was won.
}
\label{fig:defogcounts}
\end{figure}

For all variants, we observe strong win rate improvements over the control run, for both training and held-out bots.
With an auxiliary loss, we observe reductions in Q-value prediction error in the initialization training phase (\cref{fig:valid1}).
However, final win rate improvements are within the standard deviation.
Performance on the held-out bots is subject to high variance; remarkably, the gains for variants with \emph{memory} observations on the training bots can not be transferred to held-out bots. 
\cref{fig:defogcounts} illustrates unit prediction performance during two games against the the two held-out bots and reveals sensible predictions.



{
\small
\bibliography{main}
\bibliographystyle{unsrt}
}

\newpage
\appendix

\section{Action Space}
\label{sec:actions}

The actions we consider consist of switching between (or continue playing) fixed, rule-based build orders. A build order contains commands to build specific units at specific points in time, or depending on properties of the current observation, such as the unit counts per type of the player and its opponent, as well as buildings. From a reinforcement learning perspective, they can be regarded as hard-coded options that can play an entire game or be terminated at any moment. 

A build order is specialized to implement a given strategy, which corresponds to different army compositions (different types of units to build), as well as different trade-offs between short-term and long-term army strength. For instance, some build orders are specialized to assemble large armies of weaker units in the short term, while  others invest more heavily in buildings and upgrades to create stronger units in the long run. The winning probabilities of build orders against other build orders are not transitive, so the build order needs to be changed if it implements an ineffective strategy against the one chosen by the opponent. 

The action space contains 25 different build orders, each of which is specialized to a specific match-up: the game of \emph{StarCraft} has $3$ different ``races'' (Zerg, Protoss and Terran), and strategies are specialized depending on the race of the player (Zerg in our case) and its opponent.
In total, we obtain 42 distinct actions.
During a game, model outputs corresponding to build order specializations not relevant for the opponent race are ignored.

\section{Model Input}
\label{sec:features}

Our models are provided with the following features that are extracted from the game state for each model evaluation (every 5 seconds of game time).

\begin{itemize}
    \item \emph{Unit counts} are provided per-type, in disjoint channels for allied and enemy units.
    We scale the counts by approximate unit type value~\cite{synnaeve2012bayesian} and a factor of $10^{-3}$.
    We consider two variants for enemy units:
    \begin{itemize}
        \item \emph{Visible:} the currently visible enemy units.
        \item \emph{Memory:} all enemy units that have been observed since the start of the game, excluding units for which their destruction had been observed.
    \end{itemize}
    \item \emph{Resources}, i.e. minerals, gas, used supply, maximum supply are each transformed by $\log(0.2 x + 1)$.
    \item \emph{Upgrades and technologies} are marked as 1 if \emph{available} and 0 otherwise.
    \item Separately, \emph{upgrades and technologies} that are currently \emph{being researched} are marked as 1 and 0 otherwise.
    \item \emph{Game time} in minutes is transformed by $\tanh(0.1 x)$ 
    \item \emph{Build order:} index of the currently active build order
    \item \emph{Race:} index of the enemy race
    \item \emph{Map:} index of the map the game is played on
\end{itemize}

The LSTM input is a concatenation of all these features, with categorical features (race, map, build order) each represented by an 8-dimensional embedding and non-unit features (resources, upgrades and technologies) undergoing a linear projection with 8 units each.

\section{Training and Evaluation Details}
\label{sec:training}

All  games are played on the AIIDE map pool\footnote{\url{ https://skatgame.net/mburo/sc2011/rules.html}}.
During training, opponents are selected at random, and initial build orders are chosen according to a bandit algorithm.
Players can adapt between matches in short series of 25 games each.

In evaluation mode, we execute the build order with the highest value, i.e. $\argmax_a Q(o, a)$.
To reduce unnecessary back-and-forth switching, we only switch to a new build order if its value has a minimum advantage of 0.01 over the current active one. 
At the start of a game, the model receives no observations about the opponent besides its race.
Hence, we only start following the Q-function after six minutes of game time\footnote{If an opponent rush or proxy attack is detected, we allow for earlier switching.} to ensure that the selected initial build order has a sufficient effect.

During off-policy initialization, models are optimized with Adam~\cite{kingma2014adam}.
We use a learning rate of $10^{-4}$ and batch 256 games per update.
Gradients are back-propagated in time (BPTT) and truncated after 512 time steps.
$Q$-value heads are trained for build order switching points only while we compute the Huber loss for the auxiliary prediction task on every sample.
However, for each mini-batch, both losses are averaged independently, taking the total number of contributing samples into account. 
We apply a scaling factor of 10 to the loss obtained from the auxiliary task; we found no notable difference between factors of 5,10 and 20 but worse performance (in terms of value head error rates) for 1 and 100.
All models are trained for four epochs on the training corpus.

The same optimization settings are used for on-policy refinement, with the exception of a smaller batch size of 64.
The learning rate is reduced to $10^{-5}$ to improve stability during online learning.
After the initialization step above, models receive 2500 additional updates in the on-policy setting.

For testing, our model is run in evaluation, and we run three games per map against each opponent for each starting build order; players are not allowed to adapt between games to reduce the variance of the results.

\section{Auxiliary Task Performance}
\label{sec:auxperf}

\newpage
\begin{table}
   \caption{
List of opponent bots considered.
SSCAIT versions were obtained from the public SSCAIT ladder at \url{https://sscaitournament.com/index.php?action=scores};
AIIDE versions can be found at \url{https://www.cs.mun.ca/~dchurchill/starcraftaicomp/results.shtml}.
Bots denoted as ``SSCAIT*'' have been been replaced or removed online by newer versions by their respective authors. 
}
   \label{tbl:bots}
    \centering
    \small
    \begin{tabular}{lll}
    \toprule
Bot Name & Race & Version/Source \\
\midrule
AILien & Zerg & SSCAIT/AIIDE2017 \\
AIUR & Protoss & SSCAIT/AIIDE2017 \\
Arrakhammer & Zerg &  SSCAIT/AIIDE2017 \\
BananaBrain & Protoss & SSCAIT*\\
Bereaver & Protoss & SSCAIT \\
BlackCrow & Zerg & SSCAIT \\
CUNYBot & Zerg & SSCAIT \\
HannesBredberg & Terran & SSCAIT \\
HaoPan & Terran & SSCAIT \\
ICEBot & Terran & AIIDE2017 \\
Iron & Terran & SSCAIT \\
Iron & Terran & AIIDE2017 \\
Juno & Protoss & SSCAIT/AIIDE2017 \\
KillAll & Zerg & SSCAIT/AIIDE2017 \\
Killerbot & Zerg & SSCAIT \\
LetaBot & Terran & SSCAIT \\ 
LetaBot & Terran & AIIDE2017 \\
LetaBot-BBS & Terran & from Github \url{https://git.io/fxSlk} \\
LetaBot-SCVMarineRush & Terran & from Github \url{https://git.io/fxSlL} \\ 
LetaBot-SCVRush & Terran & from Github \url{https://git.io/fxSlq} \\ 
Locutus & Protoss & SSCAIT* \\ 
Locutus-2018107 & Protoss & SSCAIT \\
Matej\_Istenik & Terran & SSCAIT \\
McRave & Protoss & SSCAIT* \\
McRave-51e49b0 & Protoss & from Github \url{https://git.io/fx1ho} \\
MegaBot & Protoss & SSCAIT \\
Microwave & Zerg & SSCAIT* \\ 
NLPRBot\_CPAC & Zerg & SSCAIT/AIIDE2017 \\
NeoEdmundZerg & Zerg & SSCAIT \\
NiteKatP & Protoss & SSCAIT \\
NiteKatT & Terran & SSCAIT \\
Overkill & Zerg & SSCAIT \\
Overkill-AIIDE2016 & Zerg & AIIDE2016 \\
Overkill-AIIDE2017 & Zerg & AIIDE2017 \\
Pineapple\_Cactus & Zerg & SSCAIT \\
Prism\_Cactus & Protoss & SSCAIT \\
Proxy & Zerg & SSCAIT*  \\
Randomhammer & Protoss & SSCAIT* \\
Randomhammer & Terran & SSCAIT* \\
SkyFORKNet & Protoss & SSCAIT \\
Skynet & Protoss & SSCAIT \\
Steamhammer & Zerg & SSCAIT* \\ 
Stone & Terran & SSCAIT \\
Toothpick\_Cactus & Terran & SSCAIT \\
Tscmoo & Protoss & Provided by author \\
Tscmoo & Terran & Provided by author \\
Tscmoo & Zerg & Provided by author \\
UAlbertaBot & Protoss & SSCAIT \\
UAlbertaBot & Terran & SSCAIT \\
UAlbertaBot & Zerg & SSCAIT \\
UITTest & Protoss & SSCAIT \\
WillyT & Terran & SSCAIT* \\
WuliBot & Protoss & SSCAIT \\
Xelnaga & Protoss & AIIDE2017 \\
Ximp & Protoss & SSCAIT \\
ZZZKBot & Zerg & SSCAIT \\
Zia\_bot & Zerg & AIIDE2017 \\
    \bottomrule
   \end{tabular}
\end{table}

\end{document}